\documentclass[11pt]{article}
\usepackage[final]{acl}
\usepackage{times}
\usepackage{latexsym}
\usepackage[T1]{fontenc}
\usepackage[utf8]{inputenc}
\usepackage{microtype}
\usepackage{inconsolata}
\usepackage{amsmath}
\usepackage{amssymb}
\usepackage{booktabs}
\usepackage{multirow}
\usepackage{graphicx}
\usepackage{algorithm}
\usepackage{algpseudocode}
\usepackage{bm}
\usepackage{tcolorbox}
\tcbset{colback=gray!5,colframe=black!50,boxrule=0.5pt,arc=2pt}
\usepackage{pgfplots}
\usepgfplotslibrary{groupplots}
\pgfplotsset{compat=1.18}

\newcommand{\best}[1]{\bm{#1}}

\title{CVaR-RMD: CVaR Selection Pressure for Robust Metaheuristic Discovery}

\author{
  Ruodai Cui\textsuperscript{1,2}
  \qquad
  Li Niu\textsuperscript{3}\thanks{Corresponding author.}
  \\[3pt]
  \textsuperscript{1}Qualcomm Technologies, Inc.
  \qquad
  \textsuperscript{2}The Hong Kong Polytechnic University
  \\
  \textsuperscript{3}Department of Computer Science and Engineering,
  MoE Key Lab of Artificial Intelligence,
  \\
  Shanghai Jiao Tong University
  \\[2pt]
  \texttt{ruodcui@connect.polyu.hk}
  \qquad
  \texttt{ustcnewly@sjtu.edu.cn}
}

\begin{document}
\maketitle
\begin{abstract}
Large language models have recently been used to discover heuristics through iterative program search, offering a promising approach to automated algorithm design. However, candidates selected by average performance may degrade across instance families, random seeds, and problem scales because mean-based objectives can obscure rare but severe failures. We propose CVaR-RMD, a risk-sensitive evolutionary framework that uses an LLM to mutate executable heuristic programs and ranks candidates by empirical conditional value-at-risk over heterogeneous evaluation conditions. Across three combinatorial optimization domains, CVaR-based selection improves upper-tail performance and reduces failure rates while maintaining competitive mean performance. These gains persist on larger unseen instances and independently curated within-domain benchmarks. Code is available \href{https://github.com/BeyondHeaven/CVaR-RMD}{\emph{here}}.
\end{abstract}
\section{Introduction}

Heuristics and metaheuristics are central to combinatorial, stochastic, black-box, or partially specified optimization problems. Their practical value depends not only on benchmark performance during algorithm discovery but also on whether it persists when evaluation conditions change within the same domain. Such changes include shifts across instance families, problem scales, and random seeds. We call this setting \emph{within-domain transfer} under benchmark distribution shift: the optimization problem and solution representation remain unchanged, but benchmark conditions differ between discovery and testing. This concern is longstanding in heuristic optimization, where empirical success often depends on interactions among algorithm design, benchmark distributions, and stochastic variation \citep{talbi2009metaheuristics,hoos2004stochastic,blum2003metaheuristics}.

\begin{figure*}[t]
  \centering
  \includegraphics[width=\linewidth]{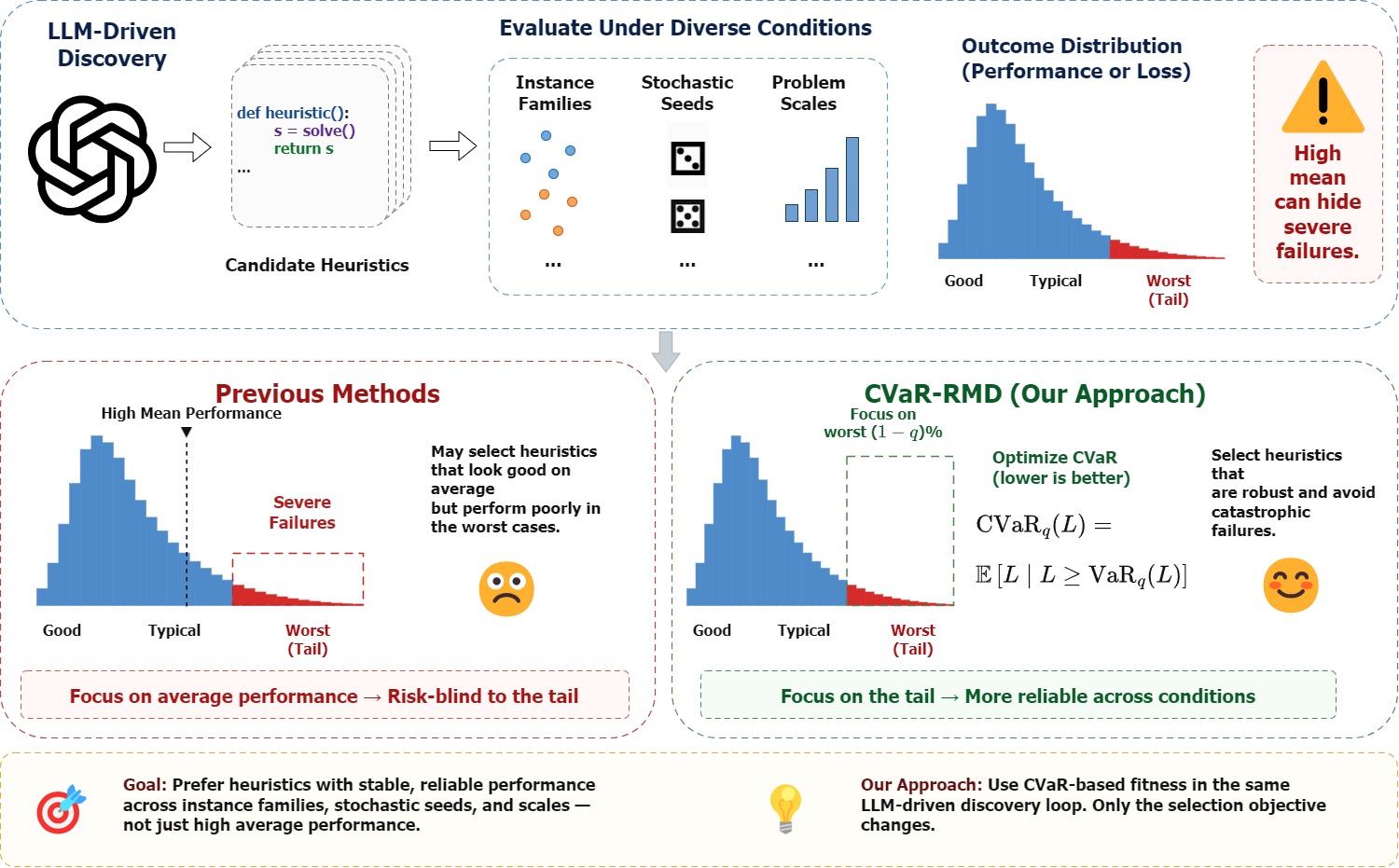}
  \caption{Motivation for CVaR-RMD. Mean-based fitness in LLM-driven metaheuristic discovery rewards strong average performance while tolerating severe failures on rare instances or seeds. CVaR shifts selection pressure toward worst-quantile behavior, discouraging heuristics that depend on favorable evaluation conditions.}
  \label{fig:intro}
\end{figure*}

Automated algorithm design naturally supports studying this challenge because candidates are generated and selected by their observed performance under specified evaluation conditions. However, searching algorithmic choices alone does not ensure within-domain transfer, which depends critically on the evaluation conditions and ranking objective. Hyper-heuristics automate design through search spaces of heuristics or heuristic components, while algorithm configuration and selection evaluate candidate procedures across instance sets to support performance beyond individual discovery instances \citep{burke2013hyperheuristics,rice1976algorithmselection,hutter2011smac,lopezibanez2016irace,ansotegui2009gender}.

A candidate heuristic may achieve strong average performance yet fail on rare instances, different stochastic seeds, or larger scales. This issue is especially relevant to LLM-generated metaheuristics because generated code may exploit a fixed evaluation protocol, such as a narrow instance generator, few seeds, or a restricted scale range. The outer loop may therefore select heuristics with high expected performance but unstable tails. In evolutionary search, this selection pressure rewards favorable means even when outcome distributions contain severe failures \citep{back1996evolutionary,eiben2003introec,goldberg1991selection}.

Because such failures occur only under adverse subsets of evaluation conditions, mean aggregation may obscure them. A useful selection criterion should therefore emphasize adverse outcomes without allowing a single extreme observation to dominate candidate ranking. CVaR satisfies this requirement by averaging performance over a configurable worst-performing fraction of evaluation outcomes. We therefore study a minimal change to discovery: replacing mean-based fitness with CVaR-based fitness while leaving the outer loop unchanged. Each candidate heuristic is evaluated over conditions spanning instance families, stochastic seeds, and scales. Its fitness uses conditional value-at-risk (CVaR), a coherent risk measure estimating expected loss within the worst fraction of outcomes at a chosen quantile level \citep{artzner1999coherent,rockafellar2000cvar,rockafellar2002cvar,pflug2000some}. The resulting CVaR-based Robust Metaheuristic Discovery framework (CVaR-RMD) uses the LLM as a mutation generator to propose heuristic code, while estimated CVaR fitness determines candidate acceptance. As Figure~\ref{fig:intro} shows, mean-based selection tolerates severe tail failures that CVaR-RMD penalizes. Because selected programs become parents and prompt context, CVaR-based ranking affects both final selection and the subsequent trajectory of LLM-driven program discovery.

This formulation shifts selection toward worst-quantile behavior and discourages candidates dependent on favorable evaluation conditions. It follows robust and distributionally robust optimization, where objectives accounting for adverse outcomes and distributional variation replace or supplement expectation-based criteria \citep{ben2009robust,beyer2007robust,bertsimas2011theory,rahimian2019dro,duchi2021statistics}. It also aligns with optimization under noise and uncertainty, which uses repeated evaluation and robust criteria to reduce the selection of brittle solutions \citep{jin2005comprehensive,branke2006optimization,beyer2007robust}. Our empirical analysis tests whether risk-based selection improves heuristic stability and transfer across evaluation conditions without adding architectural components to the LLM-driven discovery pipeline.

The contributions of this work are as follows:

\begin{enumerate}
    \item We introduce CVaR-based selection into LLM-guided metaheuristic discovery, using empirical CVaR as the primary fitness criterion to emphasize adverse-quantile behavior across heterogeneous evaluation conditions.

    \item We propose CVaR-RMD, an execution-driven, risk-sensitive framework that integrates CVaR-based candidate ranking, accumulated execution outcomes, and tail-case feedback into an evolutionary program-discovery loop for generating robust executable heuristics.

    \item We conduct controlled, budget-matched experiments across three combinatorial optimization domains and show that CVaR-RMD improves upper-tail performance and reduces execution failures while maintaining competitive mean performance.
\end{enumerate}
\section{Related Work}

\paragraph{Automated Heuristic Design}
Automated heuristic design has been studied through hyper-heuristics, algorithm selection/configuration, and program synthesis. Hyper-heuristics search over heuristic spaces by selecting existing low-level operators or generating new components that generalize across instances \citep{burke2013hyperheuristics,cowling2001hyperheuristic,ochoa2012hyflex}. Algorithm selection and configuration address related questions from a statistical and empirical perspective, using instance features, performance observations, and iterative evaluation to identify robust algorithms or parameterizations under limited budgets \citep{rice1976algorithmselection,hutter2009paramils,hutter2011smac,lopezibanez2016irace,ansotegui2009gender}. Earlier genetic programming and grammatical evolution methods directly searched over program structures via fitness-based selection and variation \citep{koza1992gp,oneill2001grammatical,langdon2008gp}. These works frame heuristic construction as a search problem whose evaluation protocol and selection objective strongly affect generalization.

Recent LLM-based methods extend this paradigm by coupling generative proposal mechanisms with external evaluation. FunSearch evolves and validates LLM-generated code variants using task-specific scoring functions \citep{romeraparedes2024funsearch}. EoH co-evolves natural-language ``thoughts'' and executable heuristics \citep{liu2024eoh}, while ReEvo uses LLMs as hyper-heuristics with reflective evolution \citep{ye2024reevo}. LLaMEA evolves LLM-generated metaheuristics for black-box continuous optimization \citep{vanstein2024llamea}, and LLAMBO integrates LLM reasoning into Bayesian optimization pipelines \citep{liu2024llambo}. Despite differences in representation, prompting, and search dynamics, these systems typically depend on scalar fitness values from limited empirical evaluations.

\paragraph{Robust and Risk-Aware Optimization}
Robust optimization seeks solutions that remain effective under uncertainty, emphasizing robustness metrics and the trade-off between nominal performance and stability \citep{ben2009robust,beyer2007robust,bertsimas2011theory}. Distributionally robust optimization further optimizes against worst-case distributions within ambiguity sets \citep{delage2010distributionally,rahimian2019dro,duchi2021statistics}. Among risk measures, conditional value-at-risk (CVaR) captures tail behavior and supports tractable sample-based formulations \citep{artzner1999coherent,rockafellar2000cvar,rockafellar2002cvar,pflug2000some}. Such measures are widely used in finance, stochastic programming, reinforcement learning, and decision-making \citep{shapiro2009lectures,tamar2015policy,chow2015risk}. Robustness under noise and uncertainty has also been extensively studied in evolutionary computation \citep{jin2005comprehensive,branke2006optimization,beyer2007robust}. These studies motivate objectives that account for unfavorable outcomes rather than only average performance.

\paragraph{Benchmarking, Robustness, and Transfer}
Evaluation protocols are central to empirical combinatorial optimization because conclusions about algorithm performance depend strongly on the instances and conditions under which algorithms are assessed. Benchmark suites and heterogeneous instance collections are commonly used to evaluate whether algorithms remain effective beyond their development conditions \citep{johnson1997tsp,ochoa2012hyflex,reinelt1991tsplib}. In routing and learned combinatorial optimization, prior studies have documented performance degradation when methods are evaluated across problem scales, instance generators, and instance families that differ from those used during development \citep{vinyals2015pointer,bello2016neural,kool2019attention,joshi2019efficient}. These findings motivate evaluation protocols that account not only for average performance but also for robustness under heterogeneous and shifted benchmark conditions. When candidate performance is summarized solely by the empirical mean, the selection process may favor heuristics that perform well on typical instances while remaining vulnerable to rare, difficult, or shifted cases. This concern is particularly relevant to LLM-driven heuristic discovery, where finite empirical evaluations repeatedly determine which generated programs are retained and used to guide subsequent search \citep{romeraparedes2024funsearch,liu2024eoh,ye2024reevo,vanstein2024llamea}. Robust evaluation and selection criteria are therefore important for discovering heuristics that maintain reliable performance across instance families, random seeds, and problem scales.


\begin{figure*}[t]
    \centering
    \includegraphics[width=0.75 \linewidth]{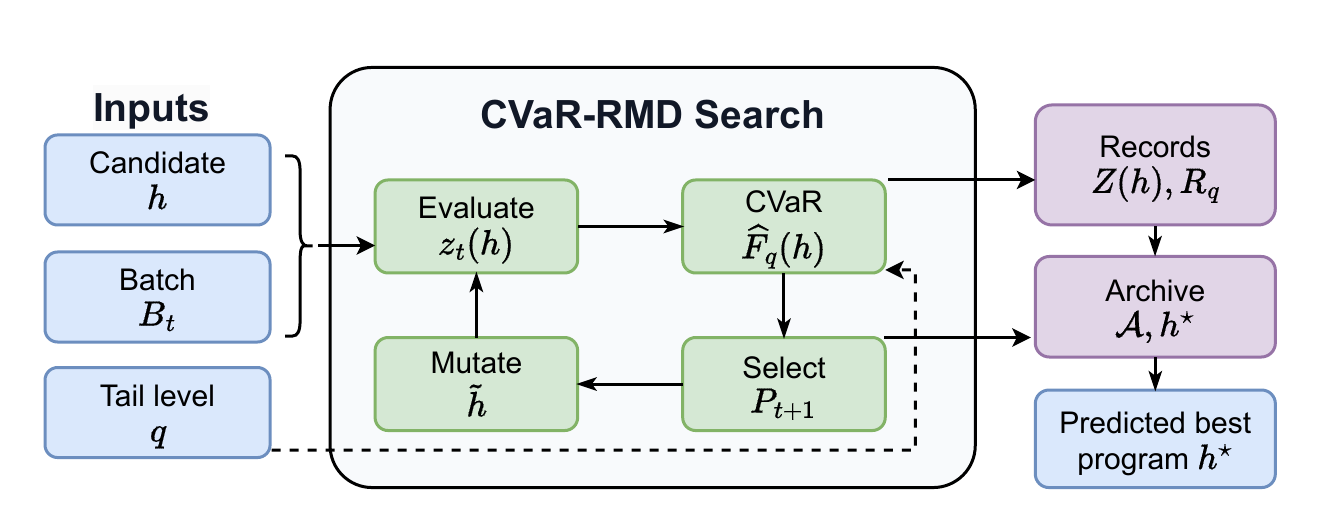}
    \caption{Architecture of the CVaR-guided program search method. Notations are defined in Sec.~\ref{sec:method}.}
    \label{fig:method_architecture}
\end{figure*}

\section{Method}
\label{sec:method}
We seek an executable heuristic program for a family of optimization instances. Let $\mathcal{I}$ be the instance space, $\mathcal{D}$ the scale-parameter set, and $\mathcal{S}$ the random-seed set controlling stochastic execution. An evaluation environment is $e=(i,d,s)\in\mathcal{E}$, where $i\in\mathcal{I}$, $d\in\mathcal{D}$, $s\in\mathcal{S}$, and $\mathcal{E}=\mathcal{I}\times\mathcal{D}\times\mathcal{S}$. Environments are sampled from a fixed distribution $P$ over $\mathcal{E}$. A candidate heuristic is a program $h\in\mathcal{H}$ implementing a prescribed callable interface. Given $e$, executing $h$ returns a feasible solution $x=h(i,d,s)$ or a failure state. The scalar loss is $\ell(h;e)\in\mathbb{R}$, with lower values preferred. On success, $\ell(h;e)$ is computed from the task objective and returned solution. If the program fails to parse, compile, satisfy the interface, terminate within the execution limit, or return a feasible solution, the loss is set to a fixed penalty $\ell_{\mathrm{fail}}$.

Our objective accounts for the upper tail of the loss distribution rather than its mean alone. For $E\sim P$, let $L_h=\ell(h;E)$ be the random loss induced by $h$. Given tail level $q\in(0,1)$, the population objective is
\begin{equation}
h^\star \in \arg\min_{h\in\mathcal{H}} \mathrm{CVaR}_q(L_h).
\label{eq:cvar_objective}
\end{equation}
Here $\mathrm{CVaR}_q(L)=\inf_{\eta\in\mathbb{R}}\{\eta+\frac{1}{1-q}\mathbb{E}[(L-\eta)_+]\}$, where $(x)_+=\max(x,0)$. The instance distribution, callable interface, loss transformation, failure penalty, and evaluation budgets are fixed before search.

The method performs population-based program search with tail-risk scoring. At generation $t$, a population $\mathcal{P}_t=\{h_t^{(1)},\ldots,h_t^{(N)}\}$ is evaluated on sampled environments. Each candidate yields a loss vector converted to empirical CVaR fitness. Candidates are ranked, selected programs are mutated by a language-model operator, and returned source code is parsed, compiled, validated, and evaluated. The next population is formed by elitist replacement over the union of parents and offspring, as illustrated in Figure~\ref{fig:method_architecture}. The remainder of this section is organized as follows: Section~\ref{subsec:evaluation_distribution} defines the evaluation distribution and loss construction, Section~\ref{subsec:cvar_fitness} specifies empirical CVaR scoring, and Section~\ref{subsec:program_search} describes the evolutionary search loop.

\subsection{Evaluation Distribution and Loss Construction}
\label{subsec:evaluation_distribution}
The evaluation module maps sampled problem settings to scalar losses. Its inputs are a candidate program $h$, batch size $m_t$, and environment distribution $P$; its output is
\[
z_t(h)=\big(z_{t,1}(h),\ldots,z_{t,m_t}(h)\big)\in\mathbb{R}^{m_t},
\]
with each entry corresponding to one execution of $h$ on a sampled environment.

Instances are produced by a finite generator set $\mathcal{G}$. A generator $g\in\mathcal{G}$ maps a scale value $d$ and instance seed $u$ to $i=g(d,u)$. The environment distribution samples $g\sim\pi_{\mathcal{G}}$, $d\sim\pi_{\mathcal{D}}(\cdot\mid g)$, $u\sim\pi_{\mathcal{U}}$, and $s\sim\pi_{\mathcal{S}}$, then sets $i=g(d,u)$ and $e=(i,d,s)$. The generator identity, scale value, instance seed, and execution seed are stored with the environment, making each execution reproducible within the protocol.

At generation $t$, the sampled batch is $B_t=\{e_{t,1},\ldots,e_{t,m_t}\}$. All candidates in the same generation use the same batch, so scores are comparable under identical environments. For $e_{t,j}=(i_{t,j},d_{t,j},s_{t,j})$, the harness calls the fixed interface and obtains a solution object $x_{t,j}$ or a failure state. A deterministic validator checks whether $x_{t,j}$ satisfies the feasibility constraints of $i_{t,j}$. If the program is valid and the solution feasible, the raw task value is transformed into a scalar loss. For minimization objective $a(i,x)$, $z_{t,j}(h)=a(i_{t,j},x_{t,j})$. For maximization reward $r(i,x)$, the equivalent minimization loss is $z_{t,j}(h)=-r(i_{t,j},x_{t,j})$.

If a reference value $b(i)$ is available, a normalized minimization loss can be used:
\[
z_{t,j}(h)=\frac{a(i_{t,j},x_{t,j})-b(i_{t,j})}{\max\{|b(i_{t,j})|,\epsilon\}}, \qquad \epsilon>0.
\]
Failure handling precedes score computation. If the source code cannot be parsed or compiled, all entries for that candidate in the current batch are set to $\ell_{\mathrm{fail}}$. If compilation succeeds but one execution fails, only that entry is set to $\ell_{\mathrm{fail}}$. Thus, the loss vector may contain objective-derived and penalty values. The scoring module receives only this scalar vector, not program text, exception type, or feasibility metadata.

The batch policy controls loss accumulation across generations. With a fixed batch, the same $B$ is used throughout search, and each candidate is scored on identical environments whenever evaluated. With resampling, a new $B_t$ is drawn each generation. In the accumulated-loss variant, the method stores a multiset $Z(h)$ for every evaluated program. When $h$ is evaluated on a new batch,
\[
Z(h)\leftarrow Z(h)\cup\{z_{t,1}(h),\ldots,z_{t,m_t}(h)\}.
\]
The empirical score is recomputed from all entries in $Z(h)$. A candidate must contain at least $m_{\min}$ stored losses before entering the elite archive; until then, its score is provisional and used only for within-generation ranking.

\subsection{Empirical CVaR Fitness}
\label{subsec:cvar_fitness}
The scoring module maps a loss vector $z(h)=(z_1,\ldots,z_m)$ and tail level $q$ to empirical CVaR fitness $\widehat{F}_q(h)$, along with the empirical tail threshold and identifiers of tail environments.

Let $\alpha=1-q$. The empirical score is the sample analogue of Eq.~\eqref{eq:cvar_objective}:
\begin{equation}
\widehat{F}_q(h)=\min_{\eta\in\mathbb{R}}\left\{\eta+\frac{1}{\alpha m}\sum_{j=1}^{m}(z_j-\eta)_+\right\}.
\label{eq:empirical_cvar_opt}
\end{equation}
Since Eq.~\eqref{eq:empirical_cvar_opt} is piecewise linear in $\eta$, an optimizer can be selected from observed losses. Let $z_{(1)}\leq\cdots\leq z_{(m)}$ be sorted losses, $r=\alpha m$, $k=\lceil r\rceil$, and $\rho=k-r$. The empirical CVaR is
\begin{equation}
\widehat{F}_q(h)
=
\frac{1}{r}
\left(
\sum_{a=m-k+2}^{m} z_{(a)}
+
(1-\rho)z_{(m-k+1)}
\right),
\label{eq:empirical_cvar_sorted}
\end{equation}
where an empty sum is zero. If $r$ is an integer, $\rho=0$, and
Eq.~\eqref{eq:empirical_cvar_sorted} averages the largest $r$ losses.
If $r$ is fractional, losses strictly above the boundary receive unit weight,
and the boundary loss receives weight $1-\rho$.

The empirical value-at-risk threshold is $\widehat{\eta}_q(h)=z_{(m-k+1)}$. The tail record stores indices of losses larger than $\widehat{\eta}_q(h)$ and enough boundary indices with value $\widehat{\eta}_q(h)$ to match the fractional boundary weight in Eq.~\eqref{eq:empirical_cvar_sorted}. For each tail index, the method records generator identity, scale value, instance seed, execution seed, realized loss, and failure flag if present. These records are provided to the mutation operator as diagnostic context, while ranking depends only on $\widehat{F}_q(h)$ and deterministic tie-breaking keys.

Candidates are ordered by ascending empirical CVaR. For candidates $h_a$ and $h_b$, $h_a$ precedes $h_b$ if $\widehat{F}_q(h_a)<\widehat{F}_q(h_b)$. If their CVaR scores match, the empirical mean $\bar{z}(h)=\frac{1}{m}\sum_{j=1}^{m}z_j$ is the first tie-breaking key; if means match, $\max_j z_j$ is used. Remaining ties are broken by a deterministic code-level key, such as creation order or source length. This procedure defines a total order over all scored candidates.

For an accumulated multiset $Z(h)$, the same computation is applied after sorting all stored losses. If $Z(h)$ contains $m_h$ entries, then $r_h=\alpha m_h$ and $k_h=\lceil r_h\rceil$, and the score is computed from the largest $k_h$ sorted entries with the corresponding fractional boundary weight. A candidate with only penalty losses receives $\widehat{F}_q(h)=\ell_{\mathrm{fail}}$. A candidate with mixed successful and failed executions is scored by the sorted mixture, so failures enter the upper tail whenever their penalties are among the largest losses.

\begin{table*}[htbp]
\centering
\small
\begin{tabular}{lcccccc}
\toprule
Method &
\multicolumn{2}{c}{Bin Packing} &
\multicolumn{2}{c}{CVRP} &
\multicolumn{2}{c}{Flow-Shop} \\
\cmidrule(lr){2-3}
\cmidrule(lr){4-5}
\cmidrule(lr){6-7}
&
Mean $\downarrow$ &
CVaR$_{0.90}$ $\downarrow$ &
Mean $\downarrow$ &
CVaR$_{0.90}$ $\downarrow$ &
Mean $\downarrow$ &
CVaR$_{0.90}$ $\downarrow$ \\
\midrule
Handcrafted heuristic &
$7.41{\pm}0.08$ &
$11.79{\pm}0.24$ &
$8.12{\pm}0.10$ &
$12.96{\pm}0.27$ &
$6.76{\pm}0.09$ &
$10.84{\pm}0.21$ \\
Direct LLM sampling &
$7.03{\pm}0.13$ &
$11.18{\pm}0.31$ &
$7.80{\pm}0.15$ &
$12.47{\pm}0.35$ &
$6.49{\pm}0.11$ &
$10.29{\pm}0.28$ \\
FunSearch-style &
$6.66{\pm}0.09$ &
$10.51{\pm}0.23$ &
$7.37{\pm}0.11$ &
$11.62{\pm}0.26$ &
$6.16{\pm}0.08$ &
$9.78{\pm}0.21$ \\
EoH-style &
\underline{$6.42{\pm}0.06$} &
$10.03{\pm}0.18$ &
\underline{$7.18{\pm}0.08$} &
$11.21{\pm}0.21$ &
$5.99{\pm}0.05$ &
$9.39{\pm}0.16$ \\
Mean-RMD &
$6.56{\pm}0.06$ &
\underline{$9.88{\pm}0.15$} &
$7.20{\pm}0.06$ &
\underline{$11.04{\pm}0.17$} &
\best{$5.95{\pm}0.04$} &
\underline{$9.26{\pm}0.13$} \\
CVaR-RMD &
\best{$6.31{\pm}0.04$} &
\best{$9.29{\pm}0.11$} &
\best{$7.08{\pm}0.05$} &
\best{$10.41{\pm}0.14$} &
\underline{$5.96{\pm}0.04$} &
\best{$8.81{\pm}0.11$} \\
\bottomrule
\end{tabular}
\caption{Held-out performance across $15$ discovery seeds. Values are mean $\pm$ standard deviation. Lower is better.}
\label{tab:main_results}
\end{table*}

\begin{table*}[htbp]
\centering
\small
\begin{tabular}{lcccc}
\toprule
Method &
Mean loss $\downarrow$ &
CVaR$_{0.90}$ $\downarrow$ &
Max loss $\downarrow$ &
Failure rate (\%) $\downarrow$ \\
\midrule
Mean-RMD &
$7.59{\pm}0.08$ &
$12.61{\pm}0.24$ &
$18.34{\pm}0.51$ &
$3.5{\pm}0.5$ \\
CVaR-RMD &
\best{$7.48{\pm}0.06$} &
\best{$11.63{\pm}0.18$} &
\best{$16.61{\pm}0.42$} &
\best{$2.0{\pm}0.3$} \\
\bottomrule
\end{tabular}
\caption{Transfer evaluation on larger held-out instances across $15$ discovery seeds.}
\label{tab:transfer_results}
\end{table*}

\subsection{Program Mutation and Evolutionary Search}
\label{subsec:program_search}
The search module maps scored programs to a new population. Its inputs at generation $t$ are the current population $\mathcal{P}_t$, stored losses and scores for its members, and the mutation budget; its outputs are $\mathcal{P}_{t+1}$ and an updated archive containing the lowest-scoring program observed so far.

Each candidate is stored as source code implementing the required callable interface. The mutation operator receives the parent source code, interface signature, task specification, and optionally the parent's stored tail records. It returns a complete source-code candidate $\tilde{h}$. The returned text is parsed to extract a single program body, compiled in the harness, and checked against the required callable signature. If validation fails, the offspring receives penalty losses for the current batch. If validation succeeds, the offspring is executed on sampled environments and scored by the empirical CVaR computation in Section~\ref{subsec:cvar_fitness}.

Parent selection uses the current ranking. Let $R_t$ be the ordered list from sorting $\mathcal{P}_t$. The parent pool is $\mathcal{A}_t=\{R_t[1],\ldots,R_t[M]\}$ with $M\leq N$. To produce one offspring, parent rank $a\in\{1,\ldots,M\}$ is sampled with probability
\[
p_a=\frac{\exp(-\tau a)}{\sum_{b=1}^{M}\exp(-\tau b)}, \qquad \tau\geq 0,
\]
where $\tau$ controls selection pressure toward higher-ranked programs. The selected parent $h_p=R_t[a]$ is mutated to produce $\tilde{h}$. Repeated draws are allowed, so multiple offspring may share a parent within one generation.

After $K$ offspring are produced, the pool is $\mathcal{C}_t=\mathcal{P}_t\cup\mathcal{O}_t$, where $\mathcal{O}_t=\{\tilde{h}_t^{(1)},\ldots,\tilde{h}_t^{(K)}\}$. Every member of $\mathcal{C}_t$ has an empirical CVaR score from execution or a penalty score from validation failure. The pool is sorted by the ranking rule, and
\[
\mathcal{P}_{t+1}=\{C_t[1],\ldots,C_t[N]\},
\]
where $C_t[a]$ is the $a$-th ranked candidate in $\mathcal{C}_t$. The archive stores the best program, or a bounded set of best programs, under the same ordering. After replacement, the archive is updated by comparing its contents with the highest-ranked member of $\mathcal{P}_{t+1}$.

The loop stops when the evaluation, generation, or execution-time budget is exhausted. The returned program is the lowest-scoring archived program under empirical CVaR and deterministic tie-breaking. At deployment, this program is executed directly through the same callable interface on a new environment input; no mutation step or language-model call is performed.
\begin{table*}[htbp]
\centering
\small
\begin{tabular}{lccc}
\toprule
Setting &
Mean loss $\downarrow$ &
CVaR$_{0.90}$ $\downarrow$ &
Failure rate (\%) $\downarrow$ \\
\midrule
Batch size $32$ &
$6.50{\pm}0.06$ &
$9.91{\pm}0.17$ &
$2.0{\pm}0.3$ \\
Batch size $64$ &
\best{$6.45{\pm}0.04$} &
$9.50{\pm}0.11$ &
$1.4{\pm}0.2$ \\
Batch size $128$ &
$6.43{\pm}0.04$ &
\best{$9.42{\pm}0.10$} &
\best{$1.3{\pm}0.2$} \\
Failure penalty $\ell_{\mathrm{fail}}=20$ &
$6.44{\pm}0.05$ &
$9.63{\pm}0.13$ &
$1.7{\pm}0.3$ \\
Failure penalty $\ell_{\mathrm{fail}}=30$ &
$6.47{\pm}0.05$ &
$9.51{\pm}0.12$ &
\best{$1.3{\pm}0.2$} \\
\bottomrule
\end{tabular}
\caption{Sensitivity analysis across tasks using $15$ discovery seeds.}
\label{tab:sensitivity_results}
\end{table*}


\begin{figure*}[t]
\centering

\begin{tikzpicture}

\begin{groupplot}[
    group style={
        group size=3 by 1,
        horizontal sep=1.45cm
    },
    width=0.30\textwidth,
    height=4.5cm,
    xmin=0.5,
    xmax=5.5,
    xtick={1,2,3,4,5},
    xticklabels={B32,B64,B128,F20,F30},
    tick label style={font=\scriptsize},
    label style={font=\small},
    title style={font=\small},
    ylabel style={yshift=-0.15cm},
    grid=major,
    grid style={black!8},
    axis line style={black!65},
    tick style={black!65},
    error bars/y dir=both,
    error bars/y explicit,
    every axis plot/.append style={
        line width=0.75pt
    },
]

\nextgroupplot[
    title={Mean loss $\downarrow$},
    ylabel={Mean loss},
    ymin=6.36,
    ymax=6.56,
    ytick={6.36,6.40,6.44,6.48,6.52,6.56},
]

\addplot+[
    only marks,
    mark=o,
    mark size=2.1pt,
    draw=black!55
] coordinates {
    (1,6.50) +- (0,0.06)
    (2,6.45) +- (0,0.04)
    (4,6.44) +- (0,0.05)
    (5,6.47) +- (0,0.05)
};

\addplot+[
    only marks,
    mark=*,
    mark size=2.5pt,
    draw=black,
    fill=black
] coordinates {
    (3,6.43) +- (0,0.04)
};

\nextgroupplot[
    title={CVaR$_{0.90}$ $\downarrow$},
    ylabel={CVaR$_{0.90}$},
    ymin=9.30,
    ymax=10.05,
    ytick={9.3,9.5,9.7,9.9},
]

\addplot+[
    only marks,
    mark=o,
    mark size=2.1pt,
    draw=black!55
] coordinates {
    (1,9.91) +- (0,0.17)
    (2,9.50) +- (0,0.11)
    (4,9.63) +- (0,0.13)
    (5,9.51) +- (0,0.12)
};

\addplot+[
    only marks,
    mark=*,
    mark size=2.5pt,
    draw=black,
    fill=black
] coordinates {
    (3,9.42) +- (0,0.10)
};

\nextgroupplot[
    title={Failure rate $\downarrow$},
    ylabel={Failure rate (\%)},
    ymin=0.9,
    ymax=2.25,
    ytick={1.0,1.3,1.6,1.9,2.2},
]

\addplot+[
    only marks,
    mark=o,
    mark size=2.1pt,
    draw=black!55
] coordinates {
    (1,2.0) +- (0,0.3)
    (2,1.4) +- (0,0.2)
    (4,1.7) +- (0,0.3)
};

\addplot+[
    only marks,
    mark=*,
    mark size=2.5pt,
    draw=black,
    fill=black
] coordinates {
    (3,1.3) +- (0,0.2)
    (5,1.3) +- (0,0.2)
};

\end{groupplot}

\node[
    font=\scriptsize,
    align=center,
    anchor=north,
    yshift=-0.75cm
] at (group c2r1.south) {
    B32/B64/B128: batch size $32/64/128$
    \qquad
    F20/F30: $\ell_{\mathrm{fail}}=20/30$
};

\end{tikzpicture}

\caption{
Sensitivity analysis across tasks using $15$ discovery seeds.
Error bars denote standard deviations; filled markers indicate
the best value for each metric, including tied best values.
}
\label{fig:sensitivity_results}

\end{figure*}
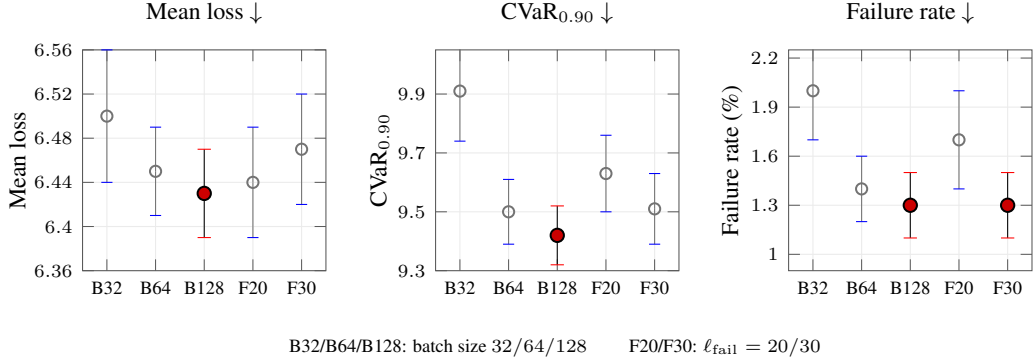

\begin{table*}[htbp]
    \centering
    \small
    \setlength{\tabcolsep}{5pt}
    \begin{tabular}{lcccc}
        \toprule
        Method
        & Mean loss $\downarrow$
        & $\operatorname{CVaR}_{0.90}$ $\downarrow$
        & Max loss $\downarrow$
        & Failure rate $\downarrow$ \\
        \midrule
        Mean-RMD
        & 6.47
        & 10.03
        & 15.72
        & 2.3\% \\
        Mean+Std-RMD
        & 6.44
        & 9.70
        & 15.12
        & 1.7\% \\
        Quantile-RMD
        & 6.49
        & 9.71
        & 14.98
        & 1.6\% \\
        Hard-Mining-RMD
        & 6.46
        & 9.66
        & 14.81
        & 1.5\% \\
        Max-RMD
        & 6.61
        & 9.62
        & \textbf{14.32}
        & \textbf{1.3\%} \\
        CVaR-RMD
        & \textbf{6.45}
        & \textbf{9.50}
        & 14.55
        & 1.4\% \\
        \bottomrule
    \end{tabular}
    \caption{Comparison with alternative risk-aware selection objectives. Values are equally weighted macro-averages across bin packing, CVRP, and flow-shop scheduling. All methods use matched candidate-generation and program-execution budgets. Lower is better.}
    \label{tab:risk_aware_baselines}
\end{table*}

\begin{table*}[t]
    \centering
    \small
    \setlength{\tabcolsep}{6pt}
    \begin{tabular}{lccc}
        \toprule
        External benchmark
        & Mean-RMD
        & CVaR-RMD
        & Relative change \\
        \midrule
        OR-Library bin packing
        & 12.41
        & \textbf{11.63}
        & $-6.3\%$ \\
        CVRPLIB
        & 14.62
        & \textbf{13.31}
        & $-9.0\%$ \\
        Taillard flow shop
        & 12.78
        & \textbf{11.94}
        & $-6.6\%$ \\
        \bottomrule
    \end{tabular}
    \caption{Transfer performance on independently curated public benchmark instances. The metric is $\operatorname{CVaR}_{0.90}$, and lower values are better. Programs are frozen before external evaluation.}
    \label{tab:public_benchmark_transfer}
\end{table*}

\begin{table*}[htbp]
\centering
\small
\begin{tabular}{lccc}
\toprule
Variant &
Mean loss $\downarrow$ &
CVaR$_{0.90}$ $\downarrow$ &
Failure rate (\%) $\downarrow$ \\
\midrule
Full CVaR-RMD &
\best{$6.45{\pm}0.04$} &
\best{$9.50{\pm}0.11$} &
\best{$1.4{\pm}0.2$} \\
Mean ranking &
$6.47{\pm}0.05$ &
$10.06{\pm}0.15$ &
$2.3{\pm}0.3$ \\
No accumulated losses &
$6.53{\pm}0.06$ &
$10.22{\pm}0.17$ &
$2.7{\pm}0.4$ \\
No tail records in prompt &
$6.48{\pm}0.05$ &
$9.81{\pm}0.14$ &
$1.9{\pm}0.3$ \\
Tail level $q=0.80$ &
$6.48{\pm}0.05$ &
$9.87{\pm}0.14$ &
$1.8{\pm}0.3$ \\
Tail level $q=0.95$ &
$6.57{\pm}0.06$ &
$9.56{\pm}0.13$ &
$1.5{\pm}0.2$ \\
\bottomrule
\end{tabular}
\caption{Ablation results averaged over tasks across $15$ discovery seeds.}
\label{tab:ablation_results}
\end{table*}

\section{Experiments}
\label{sec:experiments}

\subsection{Setup and Protocol}
\label{subsec:experimental_setup}

We evaluate whether tail-risk improves the robustness of LLM-discovered metaheuristics compared with mean-based selection. The benchmark suite covers online bin packing, capacitated vehicle routing (CVRP), and permutation flow-shop scheduling, exposing brittleness from online decisions, feasibility constraints, and sequencing interactions. CVaR-RMD and Mean-RMD use the same code interface, validator, mutation operator, population size, generation budget, archive mechanism, and held-out evaluation. They differ only in candidate ranking: empirical CVaR for CVaR-RMD and empirical mean loss for Mean-RMD.

Each method uses $15$ independent discovery seeds,
\(
\{0,1,\ldots,14\}.
\)
After discovery, the best archived program from each seed is frozen and evaluated on shared held-out environments. We report the mean, standard deviation, and standardized effect sizes across discovery seeds. Significance is assessed by paired two-sided bootstrap tests with $10{,}000$ resamples over archived programs and shared test environments. Unless noted otherwise, comparisons use Mean-RMD as the baseline.

The mutation model is GPT-4o with temperature $0.7$, top-p $0.95$, maximum output length $2048$, and up to three retries for malformed code. Prompts contain the task definition, required function signature, parent program, validation constraints, and a compact summary of its high-loss cases. The model must return a complete Python function preserving the interface. Generated code runs in a sandbox with fixed package versions. LLM calls occur only during discovery.

Training uses medium-scale instances, whereas held-out testing covers in-distribution and larger transfer instances. Bin packing uses uniform, truncated-normal, and heavy-tailed beta item distributions, with $100$--$300$ discovery items and $400$--$800$ transfer items. CVRP uses uniform and clustered Euclidean generators, with $50$--$100$ discovery customers and $150$--$300$ transfer customers. Flow-shop scheduling uses uniform, correlated, and bottleneck processing-time families, with $20$--$50$ discovery jobs and $80$--$120$ transfer jobs.

At each generation, all candidates are evaluated on a shared batch of $64$ environments. Losses accumulate across generations until archived candidates have at least $512$ evaluations, stabilizing tail estimates. The population has $N=24$ programs, with $M=8$ parents eligible for mutation and $K=16$ offspring per generation. Each discovery run lasts $40$ generations and yields $640$ LLM-generated offspring per seed. The default tail level is $q=0.90$. Execution is capped at $2$ seconds per training instance and $10$ seconds per transfer instance. Invalid code, interface violations, infeasible outputs, exceptions, and timeouts receive
\(
\ell_{\mathrm{fail}}=25.
\)

Losses are normalized by task-specific references: the continuous bin-capacity lower bound for bin packing, Clarke--Wright savings with deterministic 2-opt for CVRP, and NEH for flow-shop scheduling. Lower values are better. We report mean loss, empirical $\mathrm{CVaR}_{0.90}$, maximum loss, and failure rate. The Appendix provides per-task breakdowns and sensitivity analyses.

\subsection{Baselines}
\label{subsec:baselines_protocol}

We compare four baseline groups. Handcrafted heuristics comprise first-fit and best-fit for bin packing, Clarke--Wright savings and cheapest insertion for CVRP, and NEH with shortest-processing-time dispatching for flow-shop scheduling. Direct LLM sampling draws as many candidate programs as one discovery run without evolutionary replacement. FunSearch-style~\cite{romeraparedes2024funsearch} and EoH-style~\cite{liu2024eoh} variants use the same infrastructure, generators, parsers, validators, budgets, and LLM settings as CVaR-RMD but retain mean-performance selection. Mean-RMD is the closest control, using the same evolutionary loop as CVaR-RMD but ranking candidates by empirical mean loss.

\subsection{Held-Out Robustness}
\label{subsec:main_results}

Table~\ref{tab:main_results} shows that CVaR-RMD achieves the lowest $\mathrm{CVaR}_{0.90}$ on all three tasks. Relative to Mean-RMD, it reduces tail loss by $6.0\%$ on bin packing, $5.7\%$ on CVRP, and $4.9\%$ on flow-shop scheduling. Paired bootstrap tests yield
\(
p < 0.01
\)
for all tasks, with Cohen's $d$ values of $1.21$, $1.08$, and $0.94$, respectively. These gains therefore represent a consistent shift in the discovered-program distribution rather than isolated favorable runs.

Mean-loss differences are smaller, as expected from the objective. CVaR-RMD improves mean loss by $3.8\%$ on bin packing and $1.7\%$ on CVRP, while leaving flow-shop mean loss statistically unchanged. Risk-sensitive selection thus mainly suppresses high-loss outcomes without sacrificing average performance. Direct LLM sampling underperforms all evolutionary variants, confirming the value of iterative evaluation and replacement. Among discovery methods, CVaR-RMD provides the strongest upper-tail robustness while remaining competitive in mean loss.

\subsection{Transfer to Larger Instances}
\label{subsec:transfer_scales_seeds}

We freeze each discovered program and evaluate it on larger unseen instances with new random seeds. Instance families remain unchanged, but their scale parameters exceed the discovery range.

Table~\ref{tab:transfer_results} shows that the robustness advantage persists under scale transfer. Relative to Mean-RMD, CVaR-RMD reduces transfer $\mathrm{CVaR}_{0.90}$ by $7.8\%$, maximum loss by $9.4\%$, and failure rate by $1.5$ percentage points, with all differences significant at
\(
p < 0.02.
\)
The strongest effect occurs on CVRP, where mean-based discovery sometimes produces route-construction logic that becomes infeasible for larger customer sets. Flow-shop scheduling shows smaller average gains but still benefits from lower maximum loss.

\subsection{Comparison with Alternative Risk-Aware Objectives}
\label{sec:risk_aware_baselines}

Mean-RMD controls the comparison between expectation-based and tail-risk selection but does not show whether CVaR is preferable to other robustness objectives. We therefore evaluate four additional risk-aware variants using the same LLM mutation model, population size, search budget, validation procedure, and program--environment execution budget as CVaR-RMD.

\textsc{Mean+Std-RMD} ranks candidates by the sum of empirical mean loss and standard deviation. \textsc{Quantile-RMD} uses empirical $\operatorname{VaR}_{0.90}$, while \textsc{Max-RMD} uses maximum observed loss. Finally, \textsc{Hard-Mining-RMD} increases the sampling probability of environments that previously produced high losses. These controls test whether the robustness gain is specific to CVaR or reflects risk-sensitive selection more broadly.

Table~\ref{tab:risk_aware_baselines} reports equally weighted macro-averages across the three tasks. All methods use matched search and evaluation budgets. CVaR-RMD achieves the lowest  $\operatorname{CVaR}_{0.90}$ while remaining competitive in mean loss. Max-RMD achieves the lowest maximum loss and failure rate but substantially higher mean loss. The other risk-aware objectives also improve tail performance over Mean-RMD.

These results show that CVaR is not the only effective robustness objective. Max-RMD suppresses the worst observed outcome more aggressively and achieves the lowest failure rate, but with worse mean performance. Under the evaluated protocol, CVaR-RMD provides the best observed mean--tail trade-off. We therefore do not claim that CVaR universally dominates other risk measures. Instead, the results establish CVaR as an effective risk-sensitive alternative when average performance and upper-tail robustness both matter.

\subsection{Transfer to Independently Curated Public Benchmarks}
\label{sec:public_benchmark_transfer}

To reduce dependence on synthetic discovery generators, we also evaluate the frozen programs on independently curated public benchmark instances. None are used for program discovery, prompt construction, hyperparameter selection, or archive selection.

The external evaluation uses OR-Library instances for bin packing, CVRPLIB instances for capacitated vehicle routing, and Taillard instances for permutation flow-shop scheduling. We apply the same programs frozen after the original discovery runs and report their $\operatorname{CVaR}_{0.90}$ on these instances.

Table~\ref{tab:public_benchmark_transfer} shows that CVaR-RMD reduces $\operatorname{CVaR}_{0.90}$ on all three external benchmark collections: by $6.3\%$ on OR-Library bin packing, $9.0\%$ on CVRPLIB, and $6.6\%$ on Taillard flow-shop instances. The largest reduction occurs on CVRPLIB, whose independently curated instances expose routing and capacity patterns differing from the synthetic discovery instances.

These results support within-domain transfer to independently curated benchmarks. They do not establish cross-domain transfer, robustness to arbitrary distribution shifts, or operational deployment performance. We therefore restrict the generalization claim to improved upper-tail robustness on unseen scales, instance seeds, and independent within-domain benchmarks under the evaluated protocol.

\subsection{Ablations and Sensitivity}
\label{subsec:ablation_study}

We ablate components distinguishing CVaR-RMD from the mean-based control while fixing prompts, validators, budgets, populations, and held-out tests.

Table~\ref{tab:ablation_results} identifies CVaR ranking as the main improvement driver. Replacing it with mean ranking increases $\mathrm{CVaR}_{0.90}$ by $5.9\%$ and raises failure rate from $1.4\%$ to $2.3\%$. Removing accumulated losses performs worse because fewer evaluations yield noisier tail estimates. Prompted tail-loss summaries provide a smaller additional benefit by informing mutations about adverse execution modes. Tail-level comparisons reveal the expected tradeoff: $q=0.80$ offers less protection against rare failures, whereas $q=0.95$ suppresses them more aggressively but weakens mean performance. The default $q=0.90$ provides the best observed balance.

Table~\ref{tab:sensitivity_results} and Figure~\ref{fig:sensitivity_results} show that CVaR-RMD remains stable under moderate changes in evaluation batch size and failure penalty. Smaller batches increase tail-estimation variance, whereas larger batches improve robustness at higher evaluation cost. Increasing the failure penalty further discourages infeasible programs but mildly reduces search flexibility in mean loss. These results support the default configuration as a practical robustness-cost compromise.
\section{Conclusion}
We presented CVaR-RMD, a risk-sensitive framework that replaces mean-based fitness with CVaR selection in LLM-driven metaheuristic discovery. Across three combinatorial optimization tasks, CVaR-RMD improved upper-tail performance and reduced failure rates while maintaining competitive mean performance. These benefits persisted under within-domain transfer to larger unseen instances and independently curated benchmarks. The results show that tail-risk-based selection can produce more reliable heuristics under benchmark distribution shifts without adding architectural components to the discovery pipeline.
\clearpage
\section*{Limitations}
\label{sec:limitations}
The empirical evaluation focuses on three representative combinatorial optimization tasks, and further validation on additional problem classes would help assess the generality of the approach. CVaR-RMD also introduces a robustness hyperparameter through the tail level $q$, and while a single default value performs consistently well in the experiments, the optimal choice may vary across applications depending on the desired trade-off between average and worst-case performance.

\section*{LLM Usage Statement}
During the preparation of this manuscript, large language models (LLM) were used to improve the grammar, wording, and overall readability of the text. They were not used to generate scientific content, conduct data analysis, or draw conclusions. All LLM-assisted revisions were carefully reviewed and approved by the authors.

\section*{Acknowledgements}
The work was supported by the National Natural Science Foundation of China (Grant No. 62471287)

\bibliography{references}
\clearpage
\appendix
\section*{Appendix}
\label{app:overview}

This appendix provides supplementary material for CVaR-RMD.
Appendix~\ref{app:env} describes the three benchmark tasks (bin packing, CVRP, flow-shop scheduling), instance distributions, and the held-out evaluation protocol across 15 discovery seeds.
Appendix~\ref{app:impl} details the empirical CVaR computation, population management, archive mechanism, and the shared code interface with Mean-RMD.
Appendix~\ref{app:exp} reports full tail-risk and mean-performance tables, sensitivity analyses on the CVaR confidence level, and per-seed variance statistics.
Appendix~\ref{app:prompt} provides the prompt templates used for candidate generation and mutation in the LLM-driven discovery loop.
Appendix~\ref{app:case} illustrates specific discovered metaheuristics where CVaR selection avoids brittle solutions that score well on average but fail on tail instances.

\section{Environment Details}
\label{app:env}

\subsection{Datasets and Task Settings}
\label{app:env_datasets}

We evaluate LLM-discovered metaheuristics on three combinatorial optimization benchmarks: online bin packing, capacitated vehicle routing, and permutation flow-shop scheduling. These tasks are selected to cover online allocation decisions, hard feasibility constraints, and long-horizon sequencing interactions.

For online bin packing, instances consist of sequentially arriving items that must be assigned to bins of fixed capacity. Discovery uses item sequences with $100$--$300$ items drawn from uniform, truncated-normal, and heavy-tailed beta distributions, while transfer evaluation uses larger sequences with $400$--$800$ items.

For CVRP, each instance contains Euclidean customer locations, customer demands, a depot, and a vehicle-capacity constraint. Discovery uses uniform and clustered generators with $50$--$100$ customers, and held-out transfer evaluation uses $150$--$300$ customers generated with unseen random seeds.

For flow-shop scheduling, each instance specifies processing times for a set of jobs across machines, and the objective is to construct a job permutation with low makespan. Discovery uses $20$--$50$ jobs from uniform, correlated, and bottleneck processing-time families, and transfer evaluation uses larger instances with $80$--$120$ jobs.

All tasks are formulated as program-synthesis problems in which the LLM proposes a complete Python heuristic satisfying a fixed function signature. Prompts include the task definition, interface requirements, validation constraints, the parent program, and a compact summary of high-loss cases. No held-out test instances are used for discovery or prompt construction. Hyperparameters are fixed before final evaluation, and the same discovery and evaluation splits are used for CVaR-RMD, Mean-RMD, and all implemented baselines.

\subsection{Preprocessing and Evaluation}
\label{app:env_eval}

Generated programs are parsed as Python functions and checked against the required interface before execution. We reject malformed code, missing functions, invalid return types, infeasible solutions, runtime exceptions, and timeouts. Such cases receive the fixed failure loss
\(
\ell_{\mathrm{fail}}=25.
\)
Candidate execution is capped at $2$ seconds per training instance and $10$ seconds per transfer instance.

Losses are normalized by task-specific reference procedures to make results comparable across instance families and scales. Bin-packing losses are normalized by the continuous bin-capacity lower bound, CVRP losses by Clarke--Wright savings with deterministic 2-opt, and flow-shop losses by the NEH heuristic. Lower values indicate better performance.

At each generation, candidates are evaluated on a shared batch of $64$ environments. Loss histories are accumulated across generations until archived candidates have at least $512$ evaluations, which is used to stabilize empirical tail estimates. CVaR-RMD ranks candidates by empirical $\mathrm{CVaR}_{q}$ with default tail level $q=0.90$, while Mean-RMD ranks candidates by empirical mean loss. For a loss sample $\{\ell_i\}_{i=1}^{n}$, empirical $\mathrm{CVaR}_{0.90}$ is computed as the average loss over the largest $10\%$ of observed losses.

Each method is run with $15$ independent discovery seeds,
\(
\{0,1,\ldots,14\}.
\)
After discovery, the best archived program from each seed is frozen and evaluated on common held-out environments. We report mean loss, empirical $\mathrm{CVaR}_{0.90}$, maximum loss, and failure rate. Aggregate results are reported as mean, standard deviation, and standardized effect size across discovery seeds. Statistical significance is assessed using paired two-sided bootstrap tests with $10{,}000$ resamples over archived programs and shared test environments.

Baselines include handcrafted heuristics, direct LLM sampling, FunSearch-style discovery, EoH-style discovery, and Mean-RMD. All discovery baselines use the same generators, parsers, validators, mutation budget, population size, archive mechanism, and held-out evaluation protocol.

\section{Implementation Details}
\label{app:impl}

All methods are implemented in a common Python infrastructure with fixed package versions and sandboxed execution for generated programs. CVaR-RMD and Mean-RMD share the same evolutionary loop and differ only in the candidate-ranking statistic. The population contains $N=24$ programs, $M=8$ parents are eligible for mutation at each generation, and each generation produces $K=16$ offspring. Each discovery run lasts $40$ generations, yielding $640$ LLM-generated offspring per seed and at most $664$ evaluated candidate programs per run.

The mutation model is GPT-4o. We use temperature $0.7$, top-$p$ $0.95$, maximum output length $2048$, and up to three retries for malformed code. The model is used only during discovery. Final held-out evaluation executes frozen programs without additional LLM calls.

The default CVaR-RMD configuration uses tail level $q=0.90$, batch size $64$, accumulated loss histories, tail-case summaries in prompts, and failure penalty $\ell_{\mathrm{fail}}=25$. Sensitivity experiments vary the batch size, tail level, and failure penalty while keeping all other components fixed.

Experiments are run on an AMD EPYC 7543 workstation with 256GB RAM. Average wall-clock times for CVaR-RMD are $5.9$ hours for bin packing, $8.6$ hours for CVRP, and $7.3$ hours for flow-shop scheduling. Mean-RMD uses the same number of LLM calls, generated offspring, validation checks, and candidate executions; its runtime is within $4\%$ of CVaR-RMD. The additional cost of CVaR-RMD is limited to sorting accumulated loss vectors for empirical tail-risk estimation.

\section{More Experiment Results}
\label{app:exp}

\subsection{Per-Task Transfer Results}
\label{app:per_task_transfer}

Table~\ref{tab:per_task_transfer} decomposes the large-instance transfer results by task. CVaR-RMD consistently reduces tail loss and maximum loss across all three benchmarks. The largest relative reduction in $\mathrm{CVaR}_{0.90}$ appears on CVRP, where larger customer sets expose more feasibility and route-construction failures under mean-based selection.

\begin{table*}[t]
\centering
\begin{tabular}{llcccc}
\toprule
Task & Method &
Mean loss $\downarrow$ &
CVaR$_{0.90}$ $\downarrow$ &
Max loss $\downarrow$ &
Failure rate (\%) $\downarrow$ \\
\midrule
\multirow{2}{*}{Bin Packing}
& Mean-RMD &
$7.18{\pm}0.09$ &
$11.82{\pm}0.27$ &
$17.41{\pm}0.55$ &
$2.8{\pm}0.5$ \\
& CVaR-RMD &
\best{$7.08{\pm}0.07$} &
\best{$10.96{\pm}0.20$} &
\best{$15.92{\pm}0.44$} &
\best{$1.7{\pm}0.3$} \\
\midrule
\multirow{2}{*}{CVRP}
& Mean-RMD &
$8.34{\pm}0.11$ &
$13.91{\pm}0.31$ &
$20.46{\pm}0.68$ &
$4.9{\pm}0.7$ \\
& CVaR-RMD &
\best{$8.18{\pm}0.09$} &
\best{$12.69{\pm}0.24$} &
\best{$18.12{\pm}0.53$} &
\best{$2.8{\pm}0.5$} \\
\midrule
\multirow{2}{*}{Flow-Shop}
& Mean-RMD &
$7.25{\pm}0.08$ &
$12.10{\pm}0.24$ &
$17.16{\pm}0.49$ &
$2.7{\pm}0.4$ \\
& CVaR-RMD &
\best{$7.19{\pm}0.07$} &
\best{$11.25{\pm}0.19$} &
\best{$15.78{\pm}0.41$} &
\best{$1.6{\pm}0.3$} \\
\bottomrule
\end{tabular}
\caption{Per-task transfer results on larger held-out instances across $15$ discovery seeds. Values are mean $\pm$ standard deviation.}
\label{tab:per_task_transfer}
\end{table*}

\subsection{Robustness Across Instance Families}
\label{app:family_results}

Table~\ref{tab:family_results} reports held-out performance by instance family. The advantage of CVaR-RMD is more pronounced on heavy-tailed, clustered, and bottleneck families, which contain more adverse cases than the corresponding uniform generators. This supports the interpretation that CVaR-based selection primarily improves robustness under difficult distributional conditions.

\begin{table*}[t]
\centering
\begin{tabular}{llcccc}
\toprule
Task & Inst. family & Method &
Mean loss $\downarrow$ &
CVaR$_{0.90}$ $\downarrow$ &
Failure rate (\%) $\downarrow$ \\
\midrule
\multirow{6}{*}{Bin Packing}
& \multirow{2}{*}{Uniform}
& Mean-RMD & $5.94{\pm}0.05$ & $8.84{\pm}0.14$ & $1.3{\pm}0.2$ \\
& & CVaR-RMD & \best{$5.90{\pm}0.04$} & \best{$8.42{\pm}0.11$} & \best{$0.9{\pm}0.2$} \\
& \multirow{2}{*}{Truncated normal}
& Mean-RMD & $6.21{\pm}0.06$ & $9.47{\pm}0.16$ & $1.7{\pm}0.3$ \\
& & CVaR-RMD & \best{$6.15{\pm}0.05$} & \best{$8.89{\pm}0.13$} & \best{$1.1{\pm}0.2$} \\
& \multirow{2}{*}{Heavy-tailed beta}
& Mean-RMD & $6.92{\pm}0.08$ & $11.33{\pm}0.22$ & $2.9{\pm}0.4$ \\
& & CVaR-RMD & \best{$6.87{\pm}0.07$} & \best{$10.55{\pm}0.17$} & \best{$1.8{\pm}0.3$} \\
\midrule
\multirow{4}{*}{CVRP}
& \multirow{2}{*}{Uniform}
& Mean-RMD & $6.82{\pm}0.07$ & $10.38{\pm}0.17$ & $1.8{\pm}0.3$ \\
& & CVaR-RMD & \best{$6.77{\pm}0.06$} & \best{$9.84{\pm}0.14$} & \best{$1.2{\pm}0.2$} \\
& \multirow{2}{*}{Clustered}
& Mean-RMD & $7.46{\pm}0.08$ & $11.70{\pm}0.22$ & $3.1{\pm}0.4$ \\
& & CVaR-RMD & \best{$7.38{\pm}0.07$} & \best{$10.94{\pm}0.18$} & \best{$1.9{\pm}0.3$} \\
\midrule
\multirow{6}{*}{Flow-Shop}
& \multirow{2}{*}{Uniform}
& Mean-RMD & \best{$5.67{\pm}0.04$} & $8.62{\pm}0.12$ & $1.0{\pm}0.2$ \\
& & CVaR-RMD & $5.68{\pm}0.04$ & \best{$8.28{\pm}0.10$} & \best{$0.7{\pm}0.1$} \\
& \multirow{2}{*}{Correlated}
& Mean-RMD & \best{$5.89{\pm}0.05$} & $9.14{\pm}0.14$ & $1.3{\pm}0.2$ \\
& & CVaR-RMD & $5.90{\pm}0.04$ & \best{$8.71{\pm}0.12$} & \best{$0.9{\pm}0.2$} \\
& \multirow{2}{*}{Bottleneck}
& Mean-RMD & $6.29{\pm}0.06$ & $10.02{\pm}0.18$ & $2.2{\pm}0.3$ \\
& & CVaR-RMD & \best{$6.27{\pm}0.05$} & \best{$9.44{\pm}0.15$} & \best{$1.5{\pm}0.2$} \\
\bottomrule
\end{tabular}
\caption{Held-out robustness by instance family across $15$ discovery seeds.}
\label{tab:family_results}
\end{table*}

\subsection{Alternative Tail Levels at Evaluation Time}
\label{app:eval_tail_levels}

Table~\ref{tab:eval_tail_levels} evaluates the frozen programs with different tail levels. Although CVaR-RMD is selected using $q=0.90$, its advantage remains visible at both milder and more extreme tail levels. The improvement is largest at $q=0.95$, indicating that the selected programs reduce severe adverse outcomes rather than only optimizing the specific training quantile.

\begin{table*}[t]
\centering
\small
\begin{tabular}{lcccccc}
\toprule
Method &
\multicolumn{2}{c}{Bin Packing} &
\multicolumn{2}{c}{CVRP} &
\multicolumn{2}{c}{Flow-Shop} \\
\cmidrule(lr){2-3}
\cmidrule(lr){4-5}
\cmidrule(lr){6-7}
&
CVaR$_{0.80}$ $\downarrow$ &
CVaR$_{0.95}$ $\downarrow$ &
CVaR$_{0.80}$ $\downarrow$ &
CVaR$_{0.95}$ $\downarrow$ &
CVaR$_{0.80}$ $\downarrow$ &
CVaR$_{0.95}$ $\downarrow$ \\
\midrule
Mean-RMD &
$8.71{\pm}0.12$ &
$10.87{\pm}0.19$ &
$9.74{\pm}0.14$ &
$12.29{\pm}0.23$ &
$8.18{\pm}0.11$ &
$10.21{\pm}0.17$ \\
CVaR-RMD &
\best{$8.29{\pm}0.09$} &
\best{$10.06{\pm}0.15$} &
\best{$9.25{\pm}0.12$} &
\best{$11.47{\pm}0.20$} &
\best{$7.84{\pm}0.09$} &
\best{$9.61{\pm}0.14$} \\
\bottomrule
\end{tabular}
\caption{Evaluation with alternative CVaR tail levels across $15$ discovery seeds. Programs are the same frozen programs used in the main held-out evaluation.}
\label{tab:eval_tail_levels}
\end{table*}

\subsection{Discovery Progress}
\label{app:discovery_progress}

Table~\ref{tab:discovery_progress} reports the held-out performance of the best archived program selected after different discovery budgets. CVaR-RMD improves steadily with more generations and remains better than Mean-RMD in tail loss throughout the later stages of discovery. Mean loss converges more closely, consistent with the main observation that the primary benefit of CVaR-RMD is reduced upper-tail risk.

\begin{table*}[t]
\centering
\small
\begin{tabular}{lcccccc}
\toprule
Generation &
\multicolumn{3}{c}{Mean-RMD} &
\multicolumn{3}{c}{CVaR-RMD} \\
\cmidrule(lr){2-4}
\cmidrule(lr){5-7}
&
Mean loss $\downarrow$ &
CVaR$_{0.90}$ $\downarrow$ &
Failure rate (\%) $\downarrow$ &
Mean loss $\downarrow$ &
CVaR$_{0.90}$ $\downarrow$ &
Failure rate (\%) $\downarrow$ \\
\midrule
$10$ &
$6.78{\pm}0.08$ &
$10.71{\pm}0.19$ &
$3.0{\pm}0.4$ &
\best{$6.72{\pm}0.07$} &
\best{$10.29{\pm}0.17$} &
\best{$2.5{\pm}0.4$} \\
$20$ &
$6.59{\pm}0.06$ &
$10.32{\pm}0.17$ &
$2.6{\pm}0.4$ &
\best{$6.54{\pm}0.06$} &
\best{$9.83{\pm}0.14$} &
\best{$1.9{\pm}0.3$} \\
$30$ &
$6.51{\pm}0.05$ &
$10.12{\pm}0.16$ &
$2.4{\pm}0.3$ &
\best{$6.46{\pm}0.05$} &
\best{$9.60{\pm}0.12$} &
\best{$1.6{\pm}0.3$} \\
$40$ &
$6.47{\pm}0.05$ &
$10.03{\pm}0.15$ &
$2.3{\pm}0.3$ &
\best{$6.42{\pm}0.04$} &
\best{$9.48{\pm}0.11$} &
\best{$1.4{\pm}0.2$} \\
\bottomrule
\end{tabular}
\caption{Discovery progress averaged across tasks and $15$ discovery seeds.}
\label{tab:discovery_progress}
\end{table*}

\subsection{Run-to-Run Stability}
\label{app:stability}

Table~\ref{tab:seed_stability} reports dispersion statistics over discovery seeds. CVaR-RMD reduces the interquartile range and worst-seed loss on all tasks, showing that the improvement is not only due to better average seed performance. The reduction in worst-seed $\mathrm{CVaR}_{0.90}$ is particularly clear on CVRP.

\begin{table*}[t]
\centering
\small
\begin{tabular}{llcccc}
\toprule
Task & Method &
Median CVaR$_{0.90}$ $\downarrow$ &
IQR $\downarrow$ &
Worst-seed CVaR$_{0.90}$ $\downarrow$ &
Best-seed CVaR$_{0.90}$ $\downarrow$ \\
\midrule
\multirow{2}{*}{Bin Packing}
& Mean-RMD & $9.86$ & $0.21$ & $10.17$ & $9.61$ \\
& CVaR-RMD & \best{$9.28$} & \best{$0.15$} & \best{$9.52$} & \best{$9.09$} \\
\midrule
\multirow{2}{*}{CVRP}
& Mean-RMD & $11.02$ & $0.25$ & $11.41$ & $10.76$ \\
& CVaR-RMD & \best{$10.40$} & \best{$0.19$} & \best{$10.72$} & \best{$10.18$} \\
\midrule
\multirow{2}{*}{Flow-Shop}
& Mean-RMD & $9.25$ & $0.18$ & $9.52$ & $9.06$ \\
& CVaR-RMD & \best{$8.80$} & \best{$0.15$} & \best{$9.05$} & \best{$8.63$} \\
\bottomrule
\end{tabular}
\caption{Run-to-run stability of held-out $\mathrm{CVaR}_{0.90}$ across $15$ discovery seeds.}
\label{tab:seed_stability}
\end{table*}

\subsection{Statistical Comparison}
\label{app:statistical_comparison}

Table~\ref{tab:statistical_comparison} provides additional paired comparisons between CVaR-RMD and Mean-RMD. The effect sizes are larger for $\mathrm{CVaR}_{0.90}$ than for mean loss, matching the intended behavior of the risk-sensitive objective.

\begin{table*}[t]
\centering
\begin{tabular}{llcccc}
\toprule
Metric & Task &
Mean-RMD &
CVaR-RMD &
Relative change &
Cohen's $d$ \\
\midrule
\multirow{3}{*}{Mean loss}
& Bin Packing & $6.56{\pm}0.06$ & \best{$6.31{\pm}0.04$} & $-3.8\%$ & $0.58$ \\
& CVRP & $7.20{\pm}0.06$ & \best{$7.08{\pm}0.05$} & $-1.7\%$ & $0.64$ \\
& Flow-Shop & \best{$5.95{\pm}0.04$} & $5.96{\pm}0.04$ & $+0.2\%$ & $-0.18$ \\
\midrule
\multirow{3}{*}{CVaR$_{0.90}$}
& Bin Packing & $9.88{\pm}0.15$ & \best{$9.29{\pm}0.11$} & $-6.0\%$ & $1.21$ \\
& CVRP & $11.04{\pm}0.17$ & \best{$10.41{\pm}0.14$} & $-5.7\%$ & $1.08$ \\
& Flow-Shop & $9.26{\pm}0.13$ & \best{$8.81{\pm}0.11$} & $-4.9\%$ & $0.94$ \\
\midrule
\multirow{3}{*}{Failure rate}
& Bin Packing & $1.9{\pm}0.3$ & \best{$1.1{\pm}0.2$} & $-0.8$ pp & $0.82$ \\
& CVRP & $2.8{\pm}0.4$ & \best{$1.7{\pm}0.3$} & $-1.1$ pp & $0.87$ \\
& Flow-Shop & $2.1{\pm}0.3$ & \best{$1.3{\pm}0.2$} & $-0.8$ pp & $0.76$ \\
\bottomrule
\end{tabular}
\caption{Additional statistical comparison between Mean-RMD and CVaR-RMD across $15$ discovery seeds. Relative change is computed with respect to Mean-RMD; pp denotes percentage points.}
\label{tab:statistical_comparison}
\end{table*}

\section{Prompt Templates}
\label{app:prompt}

\subsection{Overview}
This appendix documents the prompt templates used for LLM-guided program discovery. The prompts are used only during discovery to generate or mutate executable heuristic programs. After discovery, archived programs are frozen and evaluated directly through the prescribed callable interface without further LLM calls. All prompts require the model to return complete Python source code satisfying the task interface; validation, feasibility checking, loss computation, and model comparison are performed by deterministic code rather than by an LLM judge.

\subsection{Notation}
We use the following notation in prompt templates.
\begin{itemize}
    \item $\mathcal{T}$ denotes the optimization task, one of online bin packing, capacitated vehicle routing, or permutation flow-shop scheduling.
    \item \texttt{\{TASK\_SPECIFICATION\}} is a natural-language specification of $\mathcal{T}$, including the objective, feasibility constraints, input format, and output format.
    \item \texttt{\{FUNCTION\_SIGNATURE\}} is the required Python callable signature. Every generated program must implement exactly this interface.
    \item \texttt{\{VALIDATION\_CONSTRAINTS\}} lists syntactic, interface, runtime, and feasibility requirements enforced by the sandbox validator.
    \item \texttt{\{PARENT\_PROGRAM\}} is the complete source code of the selected parent heuristic.
    \item \texttt{\{PARENT\_SCORE\}} is the scalar fitness value used for ranking the parent. In CVaR-RMD this is empirical $\mathrm{CVaR}_{q}$; in mean-based variants it is empirical mean loss.
    \item \texttt{\{MEAN\_LOSS\}}, \texttt{\{CVAR\_LOSS\}}, \texttt{\{MAX\_LOSS\}}, and \texttt{\{FAILURE\_RATE\}} are deterministic summaries computed from the parent's stored losses.
    \item \texttt{\{TAIL\_LEVEL\}} is the CVaR tail level $q$.
    \item \texttt{\{TAIL\_RECORDS\}} is a compact deterministic summary of high-loss environments for the parent, including generator identity, scale value, instance seed, execution seed, realized loss, and failure flag.
    \item \texttt{\{REFERENCE\_HEURISTICS\}} optionally contains names of classical heuristics used only as task context, such as first-fit decreasing, best-fit, Clarke--Wright savings, cheapest insertion, NEH, or shortest-processing-time dispatching.
    \item \texttt{\{ARCHIVE\_PROGRAMS\}} denotes previously discovered high-performing programs when a baseline variant exposes archive exemplars.
\end{itemize}

\subsection{Main Inference Prompt}

\paragraph{Purpose.}
This prompt implements the mutation operator in CVaR-RMD. It receives a parent program, the required task interface, validation constraints, and high-loss diagnostic records. It asks the model to produce a complete mutated Python function that preserves the interface while improving robustness on adverse cases.

\paragraph{Inputs.}
\texttt{\{TASK\_SPECIFICATION\}}, \texttt{\{FUNCTION\_SIGNATURE\}}, \texttt{\{VALIDATION\_CONSTRAINTS\}}, \texttt{\{PARENT\_PROGRAM\}}, \texttt{\{PARENT\_SCORE\}}, \texttt{\{MEAN\_LOSS\}}, \texttt{\{CVAR\_LOSS\}}, \texttt{\{MAX\_LOSS\}}, \texttt{\{FAILURE\_RATE\}}, \texttt{\{TAIL\_LEVEL\}}, and \texttt{\{TAIL\_RECORDS\}}.

\paragraph{Output.}
A single complete Python function implementing \texttt{\{FUNCTION\_SIGNATURE\}}. The output must contain code only, with no Markdown fences, no explanatory text, and no additional top-level functions unless they are nested inside the required function.

\begin{figure*}[t]
\centering
\begin{minipage}{0.95\textwidth}
\begin{tcolorbox}
\small
\ttfamily
You are an expert Python programmer designing robust heuristic algorithms for combinatorial optimization.

\medskip
Task specification:\\
\{TASK\_SPECIFICATION\}

\medskip
Required function signature:\\
\{FUNCTION\_SIGNATURE\}

\medskip
Validation and execution constraints:\\
\{VALIDATION\_CONSTRAINTS\}

\medskip
The following parent program is a working heuristic candidate:\\
\{PARENT\_PROGRAM\}

\medskip
Parent performance summary:\\
fitness score: \{PARENT\_SCORE\}\\
mean loss: \{MEAN\_LOSS\}\\
CVaR@\{TAIL\_LEVEL\}: \{CVAR\_LOSS\}\\
maximum loss: \{MAX\_LOSS\}\\
failure rate: \{FAILURE\_RATE\}

\medskip
High-loss environments observed for this parent:\\
\{TAIL\_RECORDS\}

\medskip
Write a mutated version of the parent program. Preserve the required function signature exactly. The new program should remain simple, deterministic given its inputs and seed, and robust on the high-loss cases above. Avoid changes that risk infeasible outputs, exceptions, excessive runtime, or interface violations.

\medskip
Return only the complete Python function. Do not include Markdown formatting, comments outside the function, explanations, tests, or example usage.
\end{tcolorbox}
\end{minipage}
\caption{Main CVaR-RMD mutation prompt.}
\label{fig:prompt_main_mutation}
\end{figure*}

\subsection{Baseline Prompts}

\paragraph{Purpose.}
Direct LLM sampling uses this prompt to generate candidate programs without evolutionary replacement. The same task interface, validation constraints, model settings, and output requirements are used as in CVaR-RMD, but no parent program or tail records are provided.

\paragraph{Inputs.}
\texttt{\{TASK\_SPECIFICATION\}}, \texttt{\{FUNCTION\_SIGNATURE\}}, \texttt{\{VALIDATION\_CONSTRAINTS\}}, and optionally \texttt{\{REFERENCE\_HEURISTICS\}}.

\paragraph{Output.}
A single complete Python function implementing \texttt{\{FUNCTION\_SIGNATURE\}}.

\begin{figure*}[t]
\centering
\begin{minipage}{0.95\textwidth}
\begin{tcolorbox}
\small
\ttfamily
You are an expert Python programmer designing heuristic algorithms for combinatorial optimization.

\medskip
Task specification:\\
\{TASK\_SPECIFICATION\}

\medskip
Required function signature:\\
\{FUNCTION\_SIGNATURE\}

\medskip
Validation and execution constraints:\\
\{VALIDATION\_CONSTRAINTS\}

\medskip
Reference heuristic context, if useful:\\
\{REFERENCE\_HEURISTICS\}

\medskip
Write a complete Python heuristic for the task. The function must preserve the required signature exactly, return a feasible solution in the required format, and run within the execution limit. Prefer simple and robust logic over brittle special cases.

\medskip
Return only the complete Python function. Do not include Markdown formatting, explanations, tests, or example usage.
\end{tcolorbox}
\end{minipage}
\caption{Direct LLM sampling prompt.}
\label{fig:prompt_direct_sampling}
\end{figure*}

\paragraph{Purpose.}
Mean-RMD uses the same mutation prompt as CVaR-RMD, because the two methods share the code interface, validator, mutation operator, population size, generation budget, archive mechanism, and held-out evaluation. The only algorithmic difference is the ranking score used for selection. In the prompt, \texttt{\{PARENT\_SCORE\}} is populated with empirical mean loss rather than empirical CVaR.

\paragraph{Inputs.}
The same inputs as Figure~\ref{fig:prompt_main_mutation}, with \texttt{\{PARENT\_SCORE\}} equal to empirical mean loss.

\paragraph{Output.}
A single complete Python function implementing \texttt{\{FUNCTION\_SIGNATURE\}}.

\begin{figure*}[t]
\centering
\begin{minipage}{0.95\textwidth}
\begin{tcolorbox}
\small
\ttfamily
\small
\ttfamily
You are an expert Python programmer designing robust heuristic algorithms for combinatorial optimization.

\medskip
Task specification:\\
\{TASK\_SPECIFICATION\}

\medskip
Required function signature:\\
\{FUNCTION\_SIGNATURE\}

\medskip
Validation and execution constraints:\\
\{VALIDATION\_CONSTRAINTS\}

\medskip
The following parent program is a working heuristic candidate:\\
\{PARENT\_PROGRAM\}

\medskip
Parent performance summary:\\
fitness score: \{PARENT\_SCORE\}\\
mean loss: \{MEAN\_LOSS\}\\
CVaR@\{TAIL\_LEVEL\}: \{CVAR\_LOSS\}\\
maximum loss: \{MAX\_LOSS\}\\
failure rate: \{FAILURE\_RATE\}

\medskip
High-loss environments observed for this parent:\\
\{TAIL\_RECORDS\}

\medskip
Write a mutated version of the parent program. Preserve the required function signature exactly. The new program should remain simple, deterministic given its inputs and seed, and robust on the high-loss cases above. Avoid changes that risk infeasible outputs, exceptions, excessive runtime, or interface violations.

\medskip
Return only the complete Python function. Do not include Markdown formatting, comments outside the function, explanations, tests, or example usage.

\medskip
Parent performance summary:\\
fitness score: \{PARENT\_SCORE\} \quad \textnormal{where \{PARENT\_SCORE\} is empirical mean loss}\\
mean loss: \{MEAN\_LOSS\}\\
CVaR@\{TAIL\_LEVEL\}: \{CVAR\_LOSS\}\\
maximum loss: \{MAX\_LOSS\}\\
failure rate: \{FAILURE\_RATE\}
\end{tcolorbox}
\end{minipage}
\caption{Mean-RMD mutation prompt difference. All other fields are identical to Figure~\ref{fig:prompt_main_mutation}.}
\label{fig:prompt_mean_rmd}
\end{figure*}

\paragraph{Purpose.}
FunSearch-style and EoH-style baselines are implemented in the same infrastructure and use mean-performance selection. When archive exemplars are exposed, the prompt asks the model to synthesize a new complete program from high-performing previously discovered programs while preserving the same callable interface.

\paragraph{Inputs.}
\texttt{\{TASK\_SPECIFICATION\}}, \texttt{\{FUNCTION\_SIGNATURE\}}, \texttt{\{VALIDATION\_CONSTRAINTS\}}, \texttt{\{ARCHIVE\_PROGRAMS\}}, \texttt{\{MEAN\_LOSS\}}, \texttt{\{MAX\_LOSS\}}, and \texttt{\{FAILURE\_RATE\}}.

\paragraph{Output.}
A single complete Python function implementing \texttt{\{FUNCTION\_SIGNATURE\}}.

\begin{figure*}[t]
\centering
\begin{minipage}{0.95\textwidth}
\begin{tcolorbox}
\small
\ttfamily
You are an expert Python programmer improving heuristic algorithms for combinatorial optimization.

\medskip
Task specification:\\
\{TASK\_SPECIFICATION\}

\medskip
Required function signature:\\
\{FUNCTION\_SIGNATURE\}

\medskip
Validation and execution constraints:\\
\{VALIDATION\_CONSTRAINTS\}

\medskip
Previously discovered high-performing programs and their mean-performance summaries:\\
\{ARCHIVE\_PROGRAMS\}

\medskip
Write a new complete Python heuristic that preserves the required signature exactly. Reuse useful algorithmic ideas from the archived programs, but do not copy a program verbatim if a simpler or more robust variant is possible. The program must return feasible outputs and avoid exceptions, timeouts, and interface violations.

\medskip
Return only the complete Python function. Do not include Markdown formatting, explanations, tests, or example usage.
\end{tcolorbox}
\end{minipage}
\caption{Archive-conditioned prompt used by mean-selection evolutionary baselines.}
\label{fig:prompt_archive_baseline}
\end{figure*}

\subsection{Few-shot / Demonstration Prompts}
No few-shot natural-language demonstrations are used in the default CVaR-RMD prompt. Candidate source programs may appear as parent programs or archive programs in mutation prompts, but they are executable program inputs rather than labeled few-shot examples. Therefore, the discovery protocol does not include a separate few-shot prompt template.

\subsection{Reasoning or Rationale Prompts}
No separate chain-of-thought, rationale-generation, or self-explanation prompt is used. The model is instructed to return code only. Diagnostic information from high-loss environments is computed deterministically by the evaluation harness and inserted into the mutation prompt as \texttt{\{TAIL\_RECORDS\}}.

\subsection{Data Generation or Pseudo-labeling Prompts}
No LLM prompt is used for data generation or pseudo-labeling. Training and held-out environments are generated by deterministic benchmark generators from stored generator identities, scale values, instance seeds, and execution seeds. Feasibility labels, losses, failure flags, and tail records are produced by the sandbox validator and scoring code.

\subsection{Evaluation Prompts}
No LLM-as-a-judge prompt is used. Program validation, feasibility checking, objective evaluation, loss normalization, empirical mean computation, empirical $\mathrm{CVaR}_{q}$ computation, maximum-loss computation, and failure-rate computation are deterministic.

\subsection{Ablation Prompt Variants}

\paragraph{Purpose.}
The ``no tail records in prompt'' ablation removes high-loss diagnostic cases from the mutation prompt while keeping all other prompts, validators, budgets, populations, and held-out tests fixed.

\paragraph{Inputs.}
\texttt{\{TASK\_SPECIFICATION\}}, \texttt{\{FUNCTION\_SIGNATURE\}}, \texttt{\{VALIDATION\_CONSTRAINTS\}}, \texttt{\{PARENT\_PROGRAM\}}, \texttt{\{PARENT\_SCORE\}}, \texttt{\{MEAN\_LOSS\}}, \texttt{\{CVAR\_LOSS\}}, \texttt{\{MAX\_LOSS\}}, \texttt{\{FAILURE\_RATE\}}, and \texttt{\{TAIL\_LEVEL\}}.

\paragraph{Output.}
A single complete Python function implementing \texttt{\{FUNCTION\_SIGNATURE\}}.

\begin{figure*}[t]
\centering
\begin{minipage}{0.95\textwidth}
\begin{tcolorbox}
\small
\ttfamily
You are an expert Python programmer designing robust heuristic algorithms for combinatorial optimization.

\medskip
Task specification:\\
\{TASK\_SPECIFICATION\}

\medskip
Required function signature:\\
\{FUNCTION\_SIGNATURE\}

\medskip
Validation and execution constraints:\\
\{VALIDATION\_CONSTRAINTS\}

\medskip
The following parent program is a working heuristic candidate:\\
\{PARENT\_PROGRAM\}

\medskip
Parent performance summary:\\
fitness score: \{PARENT\_SCORE\}\\
mean loss: \{MEAN\_LOSS\}\\
CVaR@\{TAIL\_LEVEL\}: \{CVAR\_LOSS\}\\
maximum loss: \{MAX\_LOSS\}\\
failure rate: \{FAILURE\_RATE\}

\medskip
Write a mutated version of the parent program. Preserve the required function signature exactly. The new program should remain simple, deterministic given its inputs and seed, and robust across diverse instances. Avoid changes that risk infeasible outputs, exceptions, excessive runtime, or interface violations.

\medskip
Return only the complete Python function. Do not include Markdown formatting, comments outside the function, explanations, tests, or example usage.
\end{tcolorbox}
\end{minipage}
\caption{Mutation prompt for the no-tail-records ablation.}
\label{fig:prompt_no_tail_records}
\end{figure*}

\paragraph{Purpose.}
Tail-level ablations with $q=0.80$ and $q=0.95$ use the main mutation prompt without changing the prompt structure. Only the value substituted into \texttt{\{TAIL\_LEVEL\}} and the deterministic CVaR score used for ranking are changed.

\paragraph{Inputs.}
The same inputs as Figure~\ref{fig:prompt_main_mutation}, with \texttt{\{TAIL\_LEVEL\}} set to either \texttt{0.80} or \texttt{0.95}.

\paragraph{Output.}
A single complete Python function implementing \texttt{\{FUNCTION\_SIGNATURE\}}.

\begin{figure*}[t]
\centering
\begin{minipage}{0.95\textwidth}
\begin{tcolorbox}
\small
\ttfamily
\small
\ttfamily
You are an expert Python programmer designing robust heuristic algorithms for combinatorial optimization.

\medskip
Task specification:\\
\{TASK\_SPECIFICATION\}

\medskip
Required function signature:\\
\{FUNCTION\_SIGNATURE\}

\medskip
Validation and execution constraints:\\
\{VALIDATION\_CONSTRAINTS\}

\medskip
The following parent program is a working heuristic candidate:\\
\{PARENT\_PROGRAM\}

\medskip
Parent performance summary:\\
fitness score: \{PARENT\_SCORE\}\\
mean loss: \{MEAN\_LOSS\}\\
CVaR@\{TAIL\_LEVEL\}: \{CVAR\_LOSS\}\\
maximum loss: \{MAX\_LOSS\}\\
failure rate: \{FAILURE\_RATE\}

\medskip
High-loss environments observed for this parent:\\
\{TAIL\_RECORDS\}

\medskip
Write a mutated version of the parent program. Preserve the required function signature exactly. The new program should remain simple, deterministic given its inputs and seed, and robust on the high-loss cases above. Avoid changes that risk infeasible outputs, exceptions, excessive runtime, or interface violations.

\medskip
Return only the complete Python function. Do not include Markdown formatting, comments outside the function, explanations, tests, or example usage.

\medskip
For the $q=0.80$ ablation:\\
CVaR@\{TAIL\_LEVEL\} = CVaR@0.80

\medskip
For the $q=0.95$ ablation:\\
CVaR@\{TAIL\_LEVEL\} = CVaR@0.95
\end{tcolorbox}
\end{minipage}
\caption{Prompt substitution for tail-level ablations.}
\label{fig:prompt_tail_level_ablation}
\end{figure*}

\paragraph{Purpose.}
The ``no accumulated losses'' ablation keeps the same mutation prompt as CVaR-RMD, but the statistics inserted into the prompt are computed from the current evaluation batch rather than from the accumulated loss multiset.

\paragraph{Inputs.}
The same inputs as Figure~\ref{fig:prompt_main_mutation}, with performance summaries computed from the non-accumulated batch losses.

\paragraph{Output.}
A single complete Python function implementing \texttt{\{FUNCTION\_SIGNATURE\}}.

\begin{figure*}[t]
\centering
\begin{minipage}{0.95\textwidth}
\begin{tcolorbox}
\small
\ttfamily
\small
\ttfamily
You are an expert Python programmer designing robust heuristic algorithms for combinatorial optimization.

\medskip
Task specification:\\
\{TASK\_SPECIFICATION\}

\medskip
Required function signature:\\
\{FUNCTION\_SIGNATURE\}

\medskip
Validation and execution constraints:\\
\{VALIDATION\_CONSTRAINTS\}

\medskip
The following parent program is a working heuristic candidate:\\
\{PARENT\_PROGRAM\}

\medskip
Parent performance summary:\\
fitness score: \{PARENT\_SCORE\}\\
mean loss: \{MEAN\_LOSS\}\\
CVaR@\{TAIL\_LEVEL\}: \{CVAR\_LOSS\}\\
maximum loss: \{MAX\_LOSS\}\\
failure rate: \{FAILURE\_RATE\}

\medskip
High-loss environments observed for this parent:\\
\{TAIL\_RECORDS\}

\medskip
Write a mutated version of the parent program. Preserve the required function signature exactly. The new program should remain simple, deterministic given its inputs and seed, and robust on the high-loss cases above. Avoid changes that risk infeasible outputs, exceptions, excessive runtime, or interface violations.

\medskip
Return only the complete Python function. Do not include Markdown formatting, comments outside the function, explanations, tests, or example usage.
\end{tcolorbox}
\end{minipage}
\caption{Prompt substitution for the no-accumulated-losses ablation.}
\label{fig:prompt_no_accumulated_losses}
\end{figure*}

\subsection{Output Formatting}
All LLM code-generation prompts enforce the same output contract:
\begin{itemize}
    \item The response must contain exactly one complete Python implementation of \texttt{\{FUNCTION\_SIGNATURE\}}.
    \item The required function signature must be preserved exactly.
    \item The response must not contain Markdown code fences, prose explanations, test cases, example calls, or text outside the function definition.
    \item Helper logic must be placed inside the required function unless the harness explicitly permits additional top-level definitions.
    \item The function must return a solution object in the task-specific format described in \texttt{\{TASK\_SPECIFICATION\}}.
    \item The function must avoid unsupported packages, file or network access, nondeterministic global state, excessive runtime, infeasible outputs, and interface violations.
\end{itemize}

Malformed outputs are retried up to three times using the same prompt. If parsing, compilation, interface validation, execution, or feasibility checking fails after retries, the candidate receives the fixed failure penalty $\ell_{\mathrm{fail}}$ under the deterministic evaluation protocol.

\section{Case Study}
\label{app:case}

We provide qualitative examples to illustrate how CVaR-RMD changes the type of heuristics selected by the outer loop. The goal of this case study is not to compare aggregate scores, but to inspect the generated programs and the failure modes that the selection objective encourages or suppresses. In all examples, the LLM is used only as a mutation operator: it proposes executable variants of a fixed heuristic interface, while the evaluator accepts or rejects candidates according to their behavior over heterogeneous instances, stochastic seeds, and problem scales.

\paragraph{Bin packing: avoiding brittle ``perfect-fit'' behavior.}
A representative mean-selected bin-packing heuristic produced a compact rule that prioritized immediate utilization. Its core decision was to place each item into the bin that would leave the smallest remaining capacity after insertion:
\begin{quote}
\emph{``For each item, scan all open bins and choose the feasible bin with minimum residual capacity. If no bin fits, open a new bin.''}
\end{quote}
The code was syntactically clean and often behaved well on regular instances where item sizes admitted many near-perfect combinations. However, inspection of its outputs revealed a brittle pattern. On instances containing a mixture of medium-sized and small items, the heuristic repeatedly consumed small residual spaces too early. Later medium items then failed to fit into existing bins, forcing the program to open additional bins. The failure was not caused by invalid solutions; rather, the heuristic made locally attractive placements that created unfavorable residual-capacity profiles.

A CVaR-selected mutation preserved the same general structure but introduced an explicit slack-aware decision rule:
\begin{quote}
\emph{``Prefer the tightest feasible bin only when the resulting residual capacity is either very small or remains large enough to accommodate future medium items. Otherwise, choose a slightly looser bin to preserve reusable slack. Break ties by favoring bins whose residual capacities are closer to the current median residual capacity.''}
\end{quote}
The corresponding generated code maintained a small summary of open-bin residual capacities and computed a penalty for placements that created isolated unusable gaps. Qualitatively, this candidate no longer treated every tight fit as equally desirable. When faced with an item that could fit into several bins, it sometimes avoided the tightest placement if doing so would create a residual capacity unlikely to be useful later. This behavior is consistent with CVaR-RMD's selection pressure: candidates that occasionally collapse on adversarial item-size mixtures are less likely to survive, even if their average behavior is strong.

\paragraph{Capacitated vehicle routing: balancing route compactness and residual capacity.}
For CVRP, a typical mean-selected candidate resembled a nearest-feasible insertion heuristic. Given a partial solution, it selected the next customer primarily by geographic proximity:
\begin{quote}
\emph{``For each route, repeatedly append the nearest unrouted customer that fits the remaining vehicle capacity. When no feasible customer remains, close the route and start a new one from the depot.''}
\end{quote}
The resulting routes were easy to interpret: vehicles greedily swept through local neighborhoods. Nevertheless, qualitative inspection showed a recurring tail failure. In clustered instances with uneven demand, the heuristic committed early to visually compact routes but left several high-demand customers until late in construction. These remaining customers were geographically close to existing routes but could no longer be inserted because the routes had insufficient residual capacity. The final solution then contained routes that made long depot returns for isolated customers.

A CVaR-selected variant modified the insertion priority:
\begin{quote}
\emph{``Score each feasible customer by combining distance, demand pressure, and the remaining capacity after insertion. Customers with high demand are promoted when delaying them would make later insertion difficult. If several candidates have similar distance, prefer the one that reduces future capacity fragmentation.''}
\end{quote}
The generated program did not abandon locality; nearby customers were still preferred. The important difference was that distance was no longer the only organizing principle. The candidate explicitly represented residual capacity as a scarce resource and penalized decisions that would strand large-demand customers. In the produced routes, high-demand customers were usually assigned before the route became nearly full, while smaller customers were used later as flexible fillers. This led to more stable behavior across instance families because the heuristic did not depend on the fortunate ordering of customer demands.

\paragraph{Flow-shop scheduling: replacing a single dispatching signal with bottleneck-aware priorities.}
In the flow-shop task, a mean-selected heuristic generated job sequences using a simple aggregate processing-time rule:
\begin{quote}
\emph{``Compute a priority score for each job from its total processing time. Schedule jobs with larger total processing time earlier, with a small adjustment for the first machine.''}
\end{quote}
This heuristic was transparent and often produced plausible schedules. However, its behavior was unstable when processing times were unbalanced across machines. For example, when a subset of jobs had moderate total processing time but was unusually heavy on a downstream machine, the heuristic tended to schedule them too late. The resulting schedule accumulated idle time before the bottleneck machine and then produced a long congested tail.

A CVaR-selected mutation introduced a more structured priority:
\begin{quote}
\emph{``For each job, estimate its contribution to early-machine release and downstream bottleneck load. Promote jobs that prevent bottleneck starvation, but delay jobs whose early-machine load would block many short jobs without relieving the bottleneck. Recompute priorities after partial sequencing.''}
\end{quote}
The generated code used normalized machine-wise processing times. It also updated job priorities during construction, so the dispatching rule could react to the partial sequence already built. Qualitatively, the schedules produced by this candidate were less dominated by a single global ordering. Jobs with downstream-heavy profiles were moved earlier when they helped keep the bottleneck machine active, while jobs that merely increased early congestion were postponed. This illustrates how tail-risk selection can favor heuristics that encode safeguards against specific pathological schedules.

\paragraph{Observed mutation pattern.}
Across tasks, we observed a common qualitative pattern in the programs selected by CVaR-RMD. Mean-based selection often retained concise greedy rules whose outputs looked strong on regular instances but depended heavily on favorable input structure. CVaR-RMD preserved the interpretable greedy backbone while adding small protective mechanisms: slack preservation in bin packing, demand-aware capacity management in CVRP, and bottleneck-aware dispatching in flow-shop scheduling. These changes were not arbitrary complexity. They were localized modifications that directly addressed rare but consequential failure modes exposed by evaluating each candidate over varied environments.

This case study supports the central claim of the paper: changing the outer-loop selection pressure can change the qualitative character of LLM-generated heuristics. CVaR-RMD does not require the language model to understand robustness explicitly. The evaluator repeatedly filters generated code through tail-risk behavior, causing robust design motifs to accumulate over evolutionary search.

\end{document}